\documentclass[letterpaper, 10 pt, conference]{ieeeconf}  
\usepackage{amsmath}
\usepackage{float}
\usepackage{graphicx}
\usepackage{multirow} 
\usepackage{diagbox}  
\usepackage{makecell}
\usepackage{algorithm}
\usepackage{algorithmic}
\usepackage{afterpage}
\usepackage{stfloats}

\IEEEoverridecommandlockouts                              

\title{\LARGE \bf
S-$\text{RRT}^*$-based Obstacle Avoidance Autonomous Motion Planner for Continuum-rigid Manipulator
}

\author{Yulin LI$^{1}$, Tetsuro Miyazaki$^{1}$, Yoshiki Yamamoto$^{1}$ and Kenji Kawashima$^{1}$
\thanks{*This work was not supported by any organization.}
\thanks{$^{1}$Department of Information Physics and Computing, 
        The University of Tokyo, 7-3-1 Hongo, Bunkyo-Ku, Tokyo, Japan
        {\tt\small kenji-kawashima@ipc.i.u-tokyo.ac.jp}}%
}

\begin{document}

\maketitle
\thispagestyle{empty}
\pagestyle{empty}

\begin{abstract}
Continuum robots are compact and flexible, making them suitable for use in the industries and in medical surgeries. 
Rapidly-exploring random trees (RRT) are a highly efficient path planning method, and its variant, S-RRT, can generate smooth feasible paths for the end-effector.
By combining RRT with inverse instantaneous kinematics (IIK), complete motion planning for the continuum arm can be achieved.
Due to the high degrees of freedom of continuum arms, the null space in IIK can be utilized for obstacle avoidance. 
In this work, we propose a novel approach that uses the S-$\text{RRT}^*$ algorithm to create paths for the continuum-rigid manipulator.
By employing IIK and null space techniques, continuous joint configurations are generated that not only track the path but also enable obstacle avoidance.
Simulation results demonstrate that our method effectively handles motion planning and obstacle avoidance while generating high-quality end-effector paths in complex environments. 
Furthermore, compared to similar IIK methods, our approach exhibits superior computation time.
\end{abstract}

\section{INTRODUCTION}
Continuum robots are highly flexible, bio-inspired systems that bridge the gap between rigid and soft robots. 
They are composed of multiple segments, each capable of bending and twisting, which allows them to adapt to complex environments \cite{cite3_overview}. 
Their unique capabilities make them widely applicable, enabling them to perform tasks such as minimally invasive surgery \cite{cite2_surgery}, aero-engine repair \cite{cite4_aerospace}, and construction \cite{cite5_constrcuting}. 
With their high degrees of freedom, passive compliance, and dexterous manipulability, continuum robots hold great potential for navigating obstacle-filled environments \cite{cite1_search}.

Motion planning is one of the most fundamental and critical problems in the field of robotics. 
Redundant robots can generate numerous solutions in joint space for the same end-effector position.
To enable redundant robots to generate feasible solutions and complete motion planning, many methods have been developed.
The sampling-based method is one of the most well-known approaches, with the rapidly exploring random tree (RRT) being a prime example for its simplicity and efficiency, along with its variant $\text{RRT}^*$ \cite{RRT_star}. 
RRT can generate a high-quality, smooth path in three dimensions, by a method known as S-RRT \cite{purning_RRT}.
RRT can also be directly deployed in the robotic configuration space (C-space) to generate feasible and continuous paths. 
Researchers have also extended this method to continuum robots \cite{RRT_continuum}.
However, the downside of the C-space RRT is that the trajectory of the end joints in the workspace(W-space) is not smooth, and the execution time is unstable. This is due to the complexity of mapping W-space obstacles into C-space. 
Another class of methods, based on inverse instantaneous kinematics (IIK), has been implemented into continuum manipulators\cite{rrt_continuum_compare}\cite{Optimization_iik}\cite{iik_Optimization}. 
By combining IIK with RRT, a feasible path can be generated in W-space for the end-effector, while IIK is used to generate the joint configuration.
This approach addresses the issue of the unsmooth end-effector path and ensures more stable computation times.
Additionally, for IIK, if the robot is redundant, the null space can be exploited to generate configurations that do not affect the end-effector's position, enabling tasks such as singularity avoidance \cite{iik_joint_1} and obstacle avoidance \cite{avoid}, for the rigid manipulator.
The use of null space for obstacle avoidance has also been implemented in continuum arms. The researchers in \cite{rrt_continuum_compare} used the null space to generate random joint configurations, enabling obstacle avoidance.
However, this method is inefficient for obstacle avoidance in low-dimensional W-space by sampling from a high-dimensional C-space, due to the complexity of C-space.
\begin{figure}[t]
    \centering
    \includegraphics[width = 1.0\linewidth]{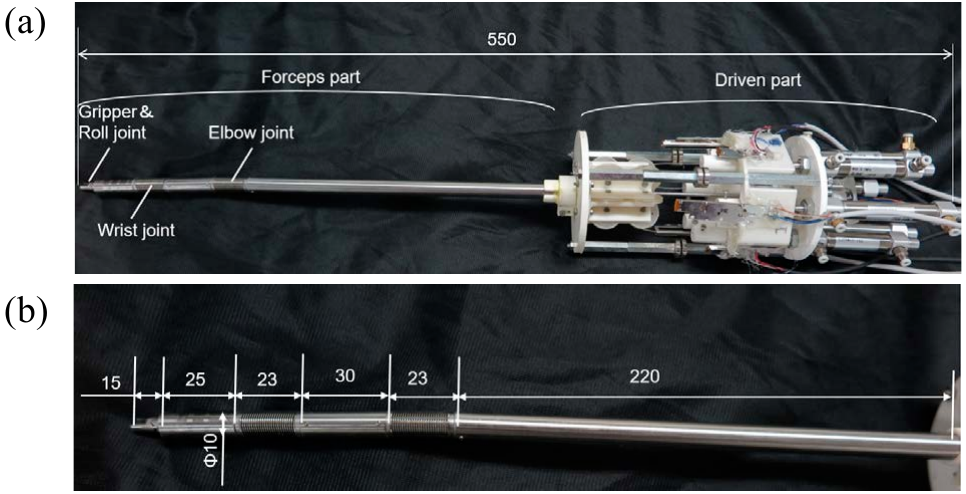}
    \caption{(a) Prototype whole part of manipulator and actuator, (b) prototype of continuum-rigid forceps manipulator.}
    \vspace{-0.2in}
    \label{model_fig}
\end{figure}
In this work, our model is a continuum-rigid manipulator, but the primary movements of the manipulator are achieved through the deformation of the continuum part, as shown in Fig.\ref{model_fig}. 
We propose a novel approach that leverages the properties of null space and combines the S-$\text{RRT}^*$ algorithm to generate high-quality end-effector paths, while enabling obstacle avoidance and autonomous motion planning for the continuum-rigid manipulator.
The main contributions of this work are: 
\begin{enumerate} 
    \item We introduce a new method for calculating the closest distance and point between an obstacle and the continuum arm. Previous research \cite{Optimization_iik} used sampled control points on the continuum arm to compute the closest distance. Our method cleverly utilized the properties of the continuum arm to calculate the closest point more efficiently, reducing computation time. 
    \item We extend the null space obstacle avoidance method \cite{avoid}, traditionally used for rigid bodies, to continuum arms. By combining this with a method for handling joint limit constraints \cite{iik_joint}, we achieve more efficient obstacle avoidance than in \cite{rrt_continuum_compare}. 
    \item We designed several different scenarios and used simulations to verify the efficiency and effectiveness of our method. In comparison with similar IIK methods \cite{Optimization_iik}, our approach demonstrates superior performance in terms of computation time. 
\end{enumerate}

\begin{figure}[b]
    \vspace{-0.1in}
    \centering
    \includegraphics[width=1.0\linewidth]{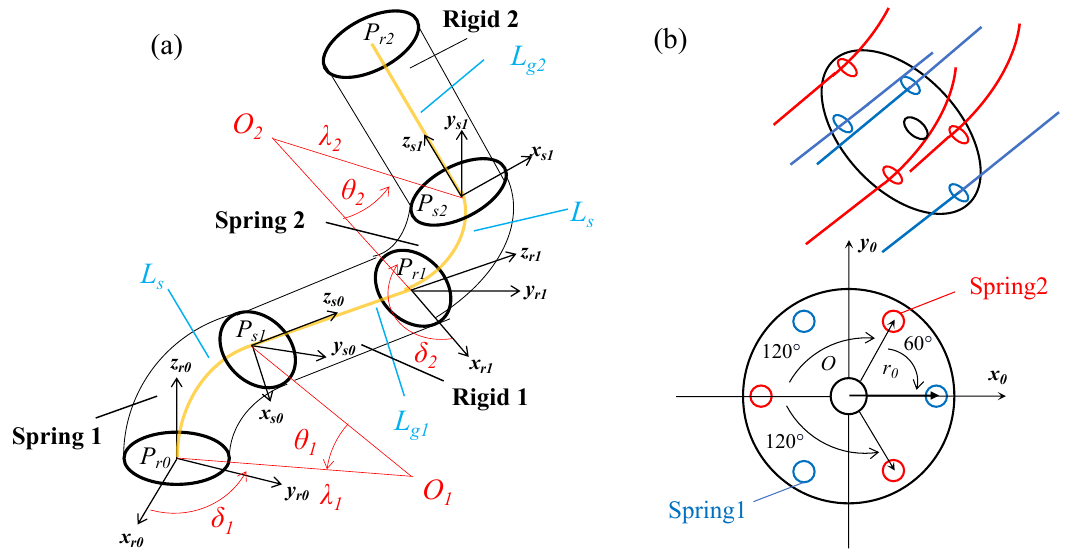}
    \caption{(a) Continuum rigid arm model by constant curvature ($\lambda_i = L_s/\theta_i$) and (b) wire arrangement.}
    \label{Model}
\end{figure}

\section{KINEMATIC MODELS}
This section discusses the kinematic models of the manipulator we use. 
The parameters of the model based on constant curvature modeling are shown in Fig.\ref{Model}(a). 
The manipulator consists of two spring-continuum arms, each with an original length of $L_s$, and two rigid arms with lengths $L_{g_1}$ and $L_{g_2}$. 
Each spring is driven by three Ni-Ti superelastic wires, which push and pull, causing the spring to deform, with the wires uniformly spaced at $120^\circ$ intervals.
The wiring diagram is illustrated in Fig.\ref{Model}(b). 
The endpoint of the bent spring $i\in (1,2)$ from the original bending plane $x_{r(i-1)}y_{r(i-1)}z_{r(i-1)}$ can be expressed as:
\begin{equation}
    \boldsymbol{p}_{si} = (L_s / \theta_{i}) [1 - \cos\theta_i, 0, \sin\theta_i]^T \label{1}
\end{equation}
where $\theta_i$ represented as the bending angle ($\theta_i \in [0, \pi]$) and $\delta_i$ represented as the wrist angle ($\delta_i \in [0, 2\pi]$). The endpoint for each rigid arm $i \in (1, 2)$ from the original plane $x_{s(i-1)}y_{s(i-1)}z_{s(i-1)}$ can be expressed as:
\begin{equation}
    \boldsymbol{p}_{ri} = L_{gi}[0, 0, 1]^T
\end{equation}
where $L_{gi}$ represented as each rigid arm length.
Given these formulations, the configuration space of this manipulator can be described using the parameters of the angles $\boldsymbol{q} = (\theta_1, \delta_1, \theta_2, \delta_2)$. 
The endpoint of each segment can be represented by a series of multiplied rotation matrices with the sum of the startpoint. 
The endpoint of each spring can be represented as:
\begin{equation}
    \boldsymbol{P}_{si} = \boldsymbol{P}_{r( i - 1)} + (\prod_{n = 0}^{i - 1} \boldsymbol{R}^{n}_{n - 1}) \boldsymbol{E}^{k\delta_i} \boldsymbol{p}_{si} \label{5}
\end{equation}
where $\boldsymbol{P}_{ri}$ is the coordinate of each endpoint of the rigid arm, $\boldsymbol{P}_{r0}$ is the original point and $\boldsymbol{R}^{i}_{i-1}$ is the combined rotation matrix, represented as:
\begin{equation}
    \boldsymbol{R}^{i}_{i - 1} = \boldsymbol{E}^{k\delta_{i}}\boldsymbol{E}^{j\theta_{i}}\boldsymbol{E}^{-k\delta_{i}}
\end{equation}
where $\boldsymbol{E}^{j\theta_i}$ represents the rotation matrix around the $y$ axis, and $\boldsymbol{E}^{k\delta_i}$ represents the rotation matrix around the $z$ axis. To clarify, $\boldsymbol{R}_{-1}^{0}$ is equal to the identity matrix. The endpoint of each rigid arm is written as follows:
\begin{equation}
    \boldsymbol{P}_{ri} = \boldsymbol{P}_{si} + (\prod_{n = 1}^{i} \boldsymbol{R}^{n}_{n - 1}) \boldsymbol{p}_{ri} \label{7}
\end{equation}
The center of each continuum arm can be expressed as:
\begin{equation}
    O_i = \boldsymbol{P}_{r( i - 1)} + (\prod_{n = 0}^{i - 1} \boldsymbol{R}^{n}_{n - 1}) \boldsymbol{E}^{k\delta_i} \boldsymbol{O}_i \label{center}
\end{equation}
where $\boldsymbol{O}_i = (L_s / \theta_i)[1 , 0, 0]^T$.

\section{METHODOLOGY}
In this section, we first discuss how to use the null space to complete obstacle avoidance. 
To achieve this, we need to calculate the closest point on the manipulator and compute its Jacobian matrix. 
Finally, we propose our strategy to achieve autonomous motion planning by combining IIK with the S-$\text{RRT}^*$ algorithm.
\subsection{Motion Planning With Inverse Instantaneous Kinematics and Obstacle Avoidance Using Null Space }
The velocity of the manipulator’s end-effector can be derived from:
\begin{equation}
\boldsymbol{\dot{p}}_e = \mathbf{J}_e \boldsymbol{\dot{q}} \label{original}
\end{equation}
where $\mathbf{J}_e$ is the Jacobian matrix for the end-effector, and $\boldsymbol{\dot{q}}$ represents the change in configuration. 
The general solution for Eq.\ref{original} is written as $\boldsymbol{\dot{q}} = \mathbf{J}_e^{\dagger}\boldsymbol{\dot{p}}_e$. 
To avoid reaching the joint limit, we introduce the weighted Jacobian matrix \cite{rrt_continuum_compare}\cite{weight}. 
Considering that the change in the end-effector’s position is very small, the new configuration $\boldsymbol{q}_{new} = \boldsymbol{q}_{old} + \boldsymbol{\dot{q}}$ can be expressed as:
\begin{equation}
\boldsymbol{q}_{new} = \boldsymbol{q}_{old} + \boldsymbol{W}^{-1/2}\mathbf{J}^{\dagger}_{we}\boldsymbol{\dot{p}}_e + (\mathbf{I} - \mathbf{J}_e^{\dagger}\mathbf{J}_e)\boldsymbol{\mu} \label{weight_jacob}
\end{equation}
where $\mathbf{J}_{we} = \mathbf{J}_e\boldsymbol{W}^{-1/2}$ is the weighted Jacobian matrix, $\boldsymbol{W}$ is the weight matrix, the operator $()^{\dagger}$ represents the Moore-Penrose pseudo-inverse, $\mathbf{I} - \mathbf{J}_e^{\dagger}\mathbf{J}_e$ is the null space and the $\boldsymbol{\mu}$ is an arbitrary vector.

To enable the manipulator to avoid obstacles, it is sufficient to assign a velocity $\boldsymbol{\dot{p}}_o$ to the point $C_o$ on the manipulator closest to the obstacle, causing it to move away from the obstacle. 
The relationship between the velocity of this point and the configuration is given by\cite{avoid}:
\begin{equation}
    \boldsymbol{\dot{p}}_o = \mathbf{J}_{C_o} \boldsymbol{\dot{q}} \label{nearest}
\end{equation}
where the $\mathbf{J}_{C_o}$ is the closest point Jacobian matrix. Combining Eq.\ref{nearest} and $\boldsymbol{\dot{q}}$, we can use arbitrary vector $\boldsymbol{\mu}$ to obtain the desired configuration that moves the closest point away from the obstacle. The arbitrary vector $\boldsymbol{\mu}$ in Eq.\ref{weight_jacob} can be represented as:
\begin{equation}
    \boldsymbol{\mu} = [\mathbf{J}_{C_o}(\mathbf{I} - \mathbf{J}_e^{\dagger}\mathbf{J}_e)]^{\dagger}(\boldsymbol{\dot{p}}_o  - \mathbf{J}_{wC_o}\mathbf{J}_{we}^{\dagger}\boldsymbol{\dot{p}}_e) \label{mu}
\end{equation}
where $\mathbf{J}_{wC_o} = \mathbf{J}_{C_o}\boldsymbol{W}^{-1/2}$ is the weighted Jacobian matrix for $C_o$. Here, we introduce $\dot{\boldsymbol{p}}_f = \dot{\boldsymbol{p}}_e + \boldsymbol{p}_{old, e} - \boldsymbol{p}_{old}$, where $\boldsymbol{p}_{old, e}$ represents the expected location of the end-effector in configuration $\boldsymbol{q}_{old}$, and $\boldsymbol{p}_{old}$ is the actual location of the end-effector. This transformation helps correct the deviation in the end-effector’s trajectory. Meanwhile, to ensure smoother transitions and to prevent the manipulator from executing obstacle avoidance when it is sufficiently far from the obstacle, we additionally introduce two coefficients, $g_h$ and $g_v$, for by combining Eq.\ref{mu} and Eq.\ref{weight_jacob}, the new configuration $\boldsymbol{q}_{new}$ can be derived as \cite{fix_path_error}\cite{fix_path_formula}:
\begin{align}
\boldsymbol{q}_{new} &= \boldsymbol{q}_{old} + \boldsymbol{W}^{-1/2}\mathbf{J}^{\dagger}_{we}\boldsymbol{\dot{p}}_f \nonumber \\
& + g_h (\mathbf{I} - \mathbf{J}_e^{\dagger}\mathbf{J}_e) [\mathbf{J}_{C_o}(\mathbf{I} - \mathbf{J}_e^{\dagger}\mathbf{J}_e)]^{\dagger}   (g_v \boldsymbol{\dot{p}}_o  - \mathbf{J}_{wC_o}\mathbf{J}_{we}^{\dagger}\boldsymbol{\dot{p}}_e) \label{final_formula}
\end{align}
The value of $g_h$ and $g_v$ is determined by the minimum distance $d$ from the manipulator to the obstacle. The formula is as follow:
\begin{equation}
    g_{h}=\left\{\begin{array}{ll}1&\quad d\leq r_{max}\\\frac{1}{2}- \frac{1}{2} \cos\left(\pi \frac{d-r_{max}}{r-r_{max}}\right)&\quad r_{max}<d<r\\0&\quad d\geq r\end{array}\right.
\end{equation}
\begin{equation}
    g_{v} = \begin{cases}
        1 & \quad d \leq r_{min} \\
        \left(\frac{d - r_{max}} {r_{max} - r_{min}}\right)^{2} & \quad r_{min} < d < r_{max} \\
        0 & \quad d \geq r_{max}
    \end{cases}
\end{equation}
where $r$, $r_{max}$, and $r_{min}$ are user-defined thresholds that reflect the extent to which the obstacle avoidance function influences the overall configuration at a given distance from the obstacle.

\subsection{Closest Point and Its Jacobian Matrix}
\begin{figure}[b]
    \vspace{-0.1in}
    \centering
    \includegraphics[width=0.9\linewidth]{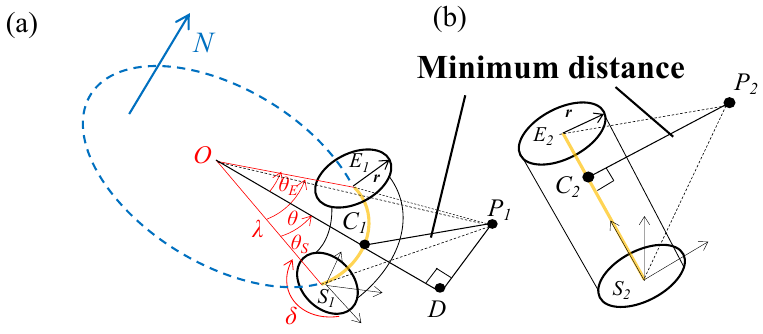}
    \caption{(a) Calculate the closest point in the continuum and (b) rigid arm. }
    \label{minimum}
\end{figure}
In our model, the closest point may be located on either the continuum or the rigid part. 
The closest point we refer to is the point on the central axis of the manipulator that is closest to the obstacle.
The closest distance is defined as the distance from the point to the center of the obstacle, minus the radius of the obstacle and the radius of the manipulator.
Assuming the obstacle can be represented as a sphere with its center at $P_i$, we only need to find the point on the central axis that is closest to the $P_i$, as described in Fig.\ref{minimum}.

To obtain the closest point $C_1$ in the continuum part, as shown in Fig.\ref{minimum}(a), the arc of the central axis must first be completed into a full circle, with the circle lying in a plane whose normal vector is $\boldsymbol{N}$. 
We first calculate the closest point in this full circle, then give a judge whether it is in the arc of the central axis. The closest point $C_s$ in the full circle can be expressed as:
\begin{equation}
    C_s = O + \lambda \frac{\overrightarrow{OD}}{|\overrightarrow{OD}|}
\end{equation}
where $D$ is the projection point of the $P_1$ on the plane of the circle, $\lambda$ is the radius of the circle and $O$ is the center of the circle. The $\overrightarrow{OD}$ can be calculated from:
\begin{equation}
    \overrightarrow{OD} = \overrightarrow{OP_1} - (\boldsymbol{N} \cdot (\overrightarrow{OP_1}))\boldsymbol{N}
\end{equation}
After obtaining $C_s$, we apply the following criteria to determine $C_1$ on the arc:
\begin{equation}
    C_1 = \begin{cases} 
        C_s & \text{if } (\theta_S + \theta_E = \theta) \\
        S_1 & \text{if } (\theta_S + \theta_E \neq \theta \text{ and } | \overrightarrow{P_1S_1} | <  | \overrightarrow{P_1E_1} |) \\
        E_1 & \text{if } (\theta_S + \theta_E \neq \theta \text{ and } | \overrightarrow{P_1S_1} | >  | \overrightarrow{P_1E_1} |)\\
        S_1 & \text{if} \quad (D = O)
    \end{cases}
\end{equation}
where $S_1$ and $E_1$ are the startpoint and the endpoint,
respectively. 
$\theta$ is the continuum part bending angle. 
$\theta_S$ and $\theta_E$ are the angles between
the vector $\overrightarrow{OC_s}$ and the vectors $\overrightarrow{OS_1}$ and $\overrightarrow{OE_1}$, respectively. 
The notation $ | \cdot | $ represents the norm, which corresponds to the length of the vector. 
This method implies that if $C_s$ lies on the arc, the $C_1$ is equal to $C_s$. If not, the $C_1$ must be either the $S_1$ or the $E_1$ of the arc. 
In an extreme case, if the projection of $P_1$, is at the center of the circle, we assign the $S_1$ as the $C_1$. 

The Jacobian matrix $\mathbf{J}_{C_o}$ of the closest point on the continuum arm can be expressed by a general solution that arbitrary point $o_1$ in the continuum arm $i$. 
We assume that the angle between the vector from the center of the arc to $o_1$ and the vector from the center to the starting point of the arc is $\theta_o$. Thus, the coordinates of $o_1$ can be written in a form similar to Eq.\ref{5}:
\begin{equation}
        \boldsymbol{P}_{so_1} = \boldsymbol{P}_{r( i - 1)} + (\prod_{n = 0}^{i - 1} \boldsymbol{R}^{n}_{n - 1}) \boldsymbol{E}^{k\delta_i} \boldsymbol{p}_{so_1}
\end{equation}
where $\boldsymbol{p}_{so_1} = (L_s / \theta_{i}) [1 - \cos (\beta \theta_i), 0, \sin (\beta \theta_i)]^T$ and $\beta = \theta_o / \theta_i$. 
The calculation method for the matrix components of $\mathbf{J}_{C_o}$ remains unchanged.

To obtain the closest point $C_2$ in the rigid part from point $P_2$, as shown in Fig. 2(b), we first consider the projection of point $P_2$ onto the line segment $S_2E_2$, denoted as $C_r$. This point may lie on the segment itself or on its extension. The vector from the rigid startpoint $S_2$ to $C_r$, denoted as $\overrightarrow{S_2C_r} = \alpha \overrightarrow{S_2E_2}$, where $\overrightarrow{S_2E_2}$ is the vector from $S_2$ to the rigid endpoint $E_2$ and $\alpha$ is a scalar, can be expressed as:
\begin{equation}
    \alpha =  \frac{\overrightarrow{S_2P_2} \cdot \overrightarrow{S_2E_2}}{| \overrightarrow{S_2E_2} | ^2}
\end{equation}
Based on this, we can determine $C_2$ using the following criteria:
\begin{equation}
    C_2 = \begin{cases} 
        S_2 & \text{if } (\alpha \leq 0) \\
        E_2 & \text{if } (\alpha \geq 1) \\
        \alpha \overrightarrow{S_2E_2} + S_2  & \text{if } (0 < \alpha < 1) \\
        
    \end{cases}
\end{equation}
This method implies that only when $0 < r < 1$, the point $C_2$ lies on the line segment $S_2E_2$. Otherwise, $C_2$ must be either the $S_2$ or the $E_2$.

For the Jacobian matrix $\mathbf{J}_{C_o}$ of the closest point on the rigid arm, we can also consider a general solution for an arbitrary point $o_2$ on the rigid arm $i$. 
We assume the distance from the startpoint of the rigid arm $i$ to $o_2$ is $L_{g_o}$. 
Thus, the coordinates of $o_2$ can be written in a form similar to Eq.\ref{7}:
\begin{equation}
        \boldsymbol{P}_{ro_2} = \boldsymbol{P}_{si} + (\prod_{n = 1}^{i} \boldsymbol{R}^{n}_{n - 1}) \boldsymbol{p}_{ro_2} 
\end{equation}
where $\boldsymbol{p}_{ro_2} = L_{g_o}[0, 0, 1]^T$ and the calculation method for $\mathbf{J}_{C_o}$ remains unchanged.
Then we can determine the closest point by comparing the minimum distance at each arm.

\subsection{S-$\text{RRT}^*$-based Autonomous Motion Planning Via Obstacle Avoidance Planner}
\vspace{-0.1in}
\begin{algorithm}[h]
    \renewcommand{\algorithmicrequire}{\textbf{Input:}}
     \renewcommand{\algorithmicensure}{\textbf{Output:}}
     \caption{S-$\text{RRT}^*$}
     \label{S_RRT}
     \begin{algorithmic}[1]
         \REQUIRE Initial point $x_{init}$; goal point $x_{goal}$; obstacle $\mathcal{O}$; search space $\mathcal{M}$; maximum number of iterations $m$
         \ENSURE A smooth path $\boldsymbol{P}_{smooth}$ from $x_{init}$ to $x_{goal}$
 
         \STATE Initial a tree $\mathcal{T}$ and add $x_{init}$ to $\mathcal{T}$;
         \WHILE{Iteration is less than $m$ \AND $x_{goal}$ not in $\mathrm{T}$}
         \STATE Generate a new point $x_{new}$ from $\mathcal{M}$;
         \IF{FreeCollision($x_{new}$)}
             \STATE Find the closest point $x_{near}$ in $\mathcal{T}$ to $x_{new}$;
             \IF{Steer($x_{near}$, $x_{new}$) is FreeCollision}
                 \STATE Add $x_{new}$ into $\mathcal{T}$ and rewire;
             \ENDIF
         \ENDIF
         \ENDWHILE
         \STATE Generate a path $\boldsymbol{P}$ from $\mathcal{T}$;
         \STATE A new path $\boldsymbol{P}_{new}$ by pruning $\boldsymbol{P}$;
         \STATE Using B-spline to smooth the $\boldsymbol{P}_{new}$ to obtain $\boldsymbol{P}_{smooth}$;
         \RETURN $\boldsymbol{P}_{smooth}$;
     \end{algorithmic}

 \end{algorithm}
 \vspace{-0.1in}
In this section, we introduce our autonomous motion planning algorithm, 
which is based on S-$\text{RRT}^*$ and incorporates the obstacle avoidance technique 
using null space as mentioned in Subsection A. The S-$\text{RRT}$ has 
been introduced in \cite{purning_RRT}, we extend it into the $\text{RRT}^{*}$ version, 
the algorithm of S-$\text{RRT}^*$ is shown in Algorithm \ref{S_RRT}. 
First, the path is generated using $\text{RRT}^*$, and then pruning is applied to obtain the minimum required points connecting the start to the end. 
Finally, a smooth path is generated using B-spline interpolation. Our autonomous motion planning algorithm is shown in Algorithm \ref{S_RRT_motion_planning}, 
we first obtain the end-effector path by using S-$\text{RRT}^{*}$, then calculate the configuration for the correspond end-effector path. When calculating the new configuration, 
unlike previous works\cite{avoid}, 
we first set $\boldsymbol{\mu} = 0$ in Eq.\ref{weight_jacob} and compute a new configuration to obtain the velocity of the closest point. 
If the closest point is not moving toward the obstacle, we directly add this configuration to our configuration array.
Only when the closest point is moving toward the obstacle do we use Eq.\ref{final_formula} to compute a new configuration that moves the closest point away from the obstacle.
The velocity $\boldsymbol{\dot{p}}_0$ is calculated as $-k (\overrightarrow{C_oO} / |\overrightarrow{C_oO}|)$, where $\overrightarrow{C_oO}$ is the vector from the closest point on the manipulator to the center of the closest obstacle and $k$ is user defined positive constant.
\vspace{-0.1in}
\begin{algorithm}[h]
    \renewcommand{\algorithmicrequire}{\textbf{Input:}}
    \renewcommand{\algorithmicensure}{\textbf{Output:}}
    \caption{S-$\text{RRT}^*$-bsaed Obstacle Avoidance Autonomous Motion Planner}
    \label{S_RRT_motion_planning}
    \begin{algorithmic}[1]
        \REQUIRE Initial configuration $q_{init}$; initial startpoint $x_{init}$; endpoint $x_{goal}$; obstacles $\mathcal{O}$; search space $\mathcal{M}$; maximum number of iterations $m$;
        \ENSURE Path for the end-effector $\boldsymbol{P}_e$ and sequence obstacles avoidance configurations $\mathcal{Q}$

        \STATE Obtain $\boldsymbol{P}_{e}$ for the end-effector from S-$\text{RRT}^*$($x_{init}$, $x_{goal}$, $\mathcal{O}$, $\mathcal{M}$, $m$);
        \STATE Initial $\mathcal{Q}$ ,add $\boldsymbol{q}_{init}$ into $\mathcal{Q}$ and let $i = 0$;
        \FOR{$i + 1$ less than length of $\boldsymbol{P}_e$}
           \STATE Velocity $\dot{\boldsymbol{p}}_e$ eqaul to $\boldsymbol{P}_e$[$i + 1$] - $\boldsymbol{P}_e$[$i$];
           \STATE Find the closest point $C_o$ to $\mathcal{O}$ in manipulator, obtain the vector from the closest $\mathcal{O}$ center $\overrightarrow{C_oO}$;
           \STATE Calculate the end-effector and $C_o$ Jacobian matrix;
           \STATE Calculate a temporary configuration $\boldsymbol{q}_{tem}$ from Eq.\ref{weight_jacob}, where $\boldsymbol{\mu} = 0$;
           \STATE Calculate the $C_o$ velocity $\boldsymbol{\dot{p}}_{o, new}$ from Eq.\ref{nearest};
           \IF{projection of $\boldsymbol{\dot{p}}_{o, new}$ on $\overrightarrow{C_oO}$ is positive \AND $C_o$ is not located in end-effector}
               \STATE Assign an escape velocity $-k(\overrightarrow{C_oO}/|\overrightarrow{C_oO}|)$ for $\boldsymbol{\dot{p}}_0$ in Eq.\ref{final_formula}, then recalculation $\boldsymbol{q}_{tem}$ based on Eq.\ref{final_formula};
           \ENDIF
           \STATE Add $\boldsymbol{q}_{tem}$ into $\mathcal{Q}$;
           \STATE $i = i + 1$;
        \ENDFOR
        \RETURN $\boldsymbol{P}_e$ and $\mathcal{Q}$;
    \end{algorithmic}

\end{algorithm}
\vspace{-0.1in}

\section{EXPERIMENT}

In this section, we verify the effectiveness of our S-$\text{RRT}^*$-based path planning obstacle avoidance method.
We demonstrate that our method can successfully complete motion planning in various scenarios, producing high-quality end-effector trajectories with faster computation times compared to previous methods. 
The model used in this study is a four-segment cable-driven continuum-rigid manipulator, as shown in Fig.\ref{Model}. The simulations were conducted in Python, with parameters derived from our model. All simulations were executed on a computer equipped with an Apple M3 chip and 16GB of RAM.

\subsection{Motion Planning Via Fixed Path}
\begin{figure}[t]
    \vspace{0.055in}
    \centering
    \includegraphics[width=0.9\linewidth]{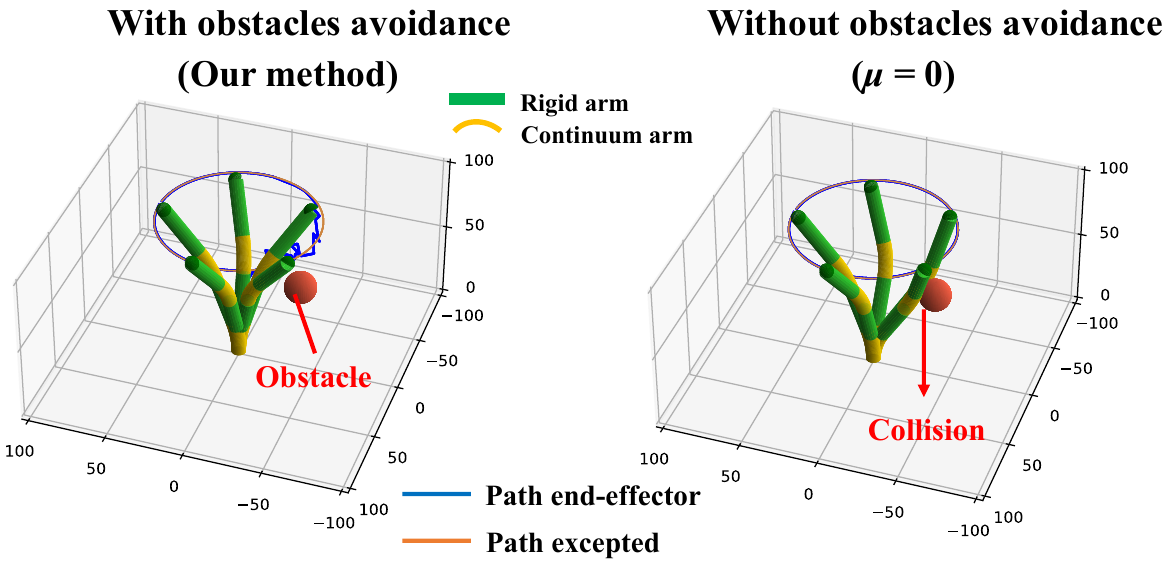}
    \caption{Motion planning with and without obstacle avoidance for a fixed circle path. The $\boldsymbol{\mu}$ refer to the $\boldsymbol{\mu}$ in the Eq.\ref{weight_jacob}.}
    \label{Simulation_fix}
    \vspace{-0.22in}
\end{figure}

\begin{figure}[t]
    \vspace{0.055in}
    \centering
    \includegraphics[width=1\linewidth]{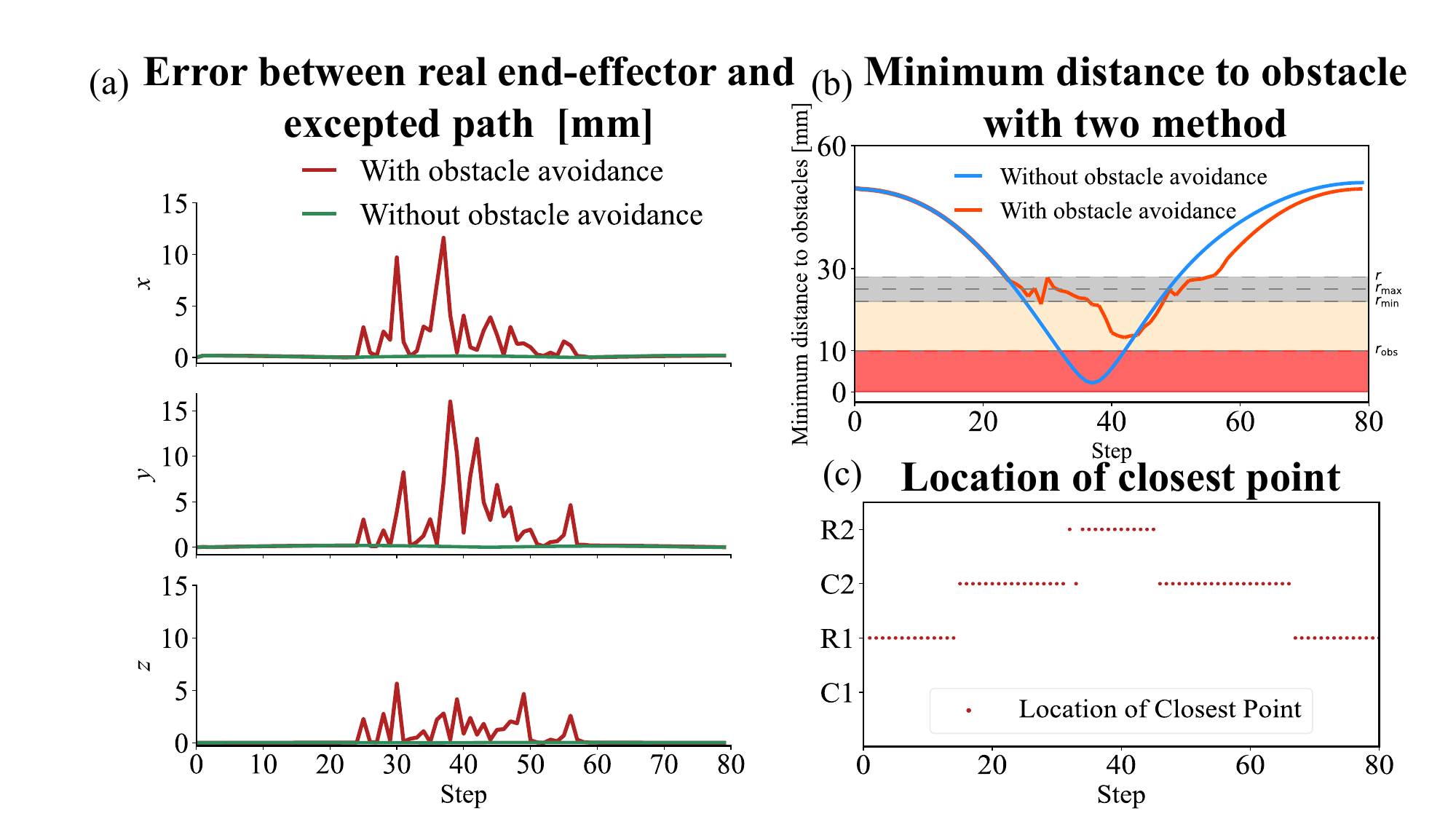}
    \caption{Quantitative analysis of fixed path motion planning with two planners
    (a) Error between the end-effector location and the expected path, 
    (b) Minimum distance to the obstacle at each step for both planners, 
    (c) Location of the closest point at each step for the obstacle avoidance planner. (C refer to continuum arm, R refer to rigid arm and numbers refer to their index.)}
    \label{Ana_fix}
    \vspace{-0.22in}
\end{figure}

This section defines a fixed circular path to validate the performance of using null space to achieve obstacle avoidance, 
ensuring that the closest point, whether in the continuum or rigid arm, 
moves away from the obstacle. Additionally, 
we analyze the error between the end-effector’s trajectory and the expected path, the change of closet point, and the minimum distance to the obstacle in each step, as shown in Fig.\ref{Simulation_fix} and Fig.\ref{Ana_fix}. 
The initial configuration is $[\pi / 9, 0, \pi/9, 0]$, with the startpoint at $[51, 0, 101]$. The fixed circle’s center is located at $[0, 0, 101]$ with a radius of 51 mm. 
The $r$, $r_{max}$ and $r_{min}$ are set to 28 mm, 25 mm and 22 mm.
The obstacle is placed at $[-40, 0, 60]$ with a radius of 10 mm, and the escape velocity $-k (\overrightarrow{C_oO} / | \overrightarrow{C_oO} |)$, where $k$ is set to 6.

The motion planning process with and without obstacle avoidance, where $\boldsymbol{\mu}$ in Eq.\ref{weight_jacob} is set to 0, is shown in Fig.\ref{Simulation_fix}. 
We also quantitatively analyze the error between the end-effector trajectory and the actual trajectory during the motion planning process for both methods, as shown in Fig.\ref{Ana_fix}(a).
Both methods demonstrate high accuracy in tracking the given path. For the planner without obstacle avoidance, the error is closest to 0. Even though the obstacle avoidance planner generates some deviation from the given path when attempting to avoid collisions with the obstacle, it eventually returns to the designated path. 
These deviations are the trade-off for incorporating the obstacle avoidance function, as shown in Fig.\ref{Ana_fix}(b).
Without the obstacle avoidance planner, the manipulator collides with the obstacle. However, with the obstacle avoidance planner, once the minimum distance to the obstacle is less than $r$, the planner activates and successfully guides the manipulator away from the obstacle, preventing a collision.
We also validate the effectiveness of obstacle avoidance in the continuum part. Fig.\ref{Ana_fix}(c) shows the changes in the closest point on the manipulator when the obstacle avoidance planner is enabled. After the minimum distance to the obstacle falls below $r$, the closest point is primarily located on the second continuum segment and the second rigid arm.

\subsection{Motion Planning Via Various Environments}
\begin{figure}[b]
    \vspace{-0.1in}
    \centering
    \includegraphics[width=0.8\linewidth]{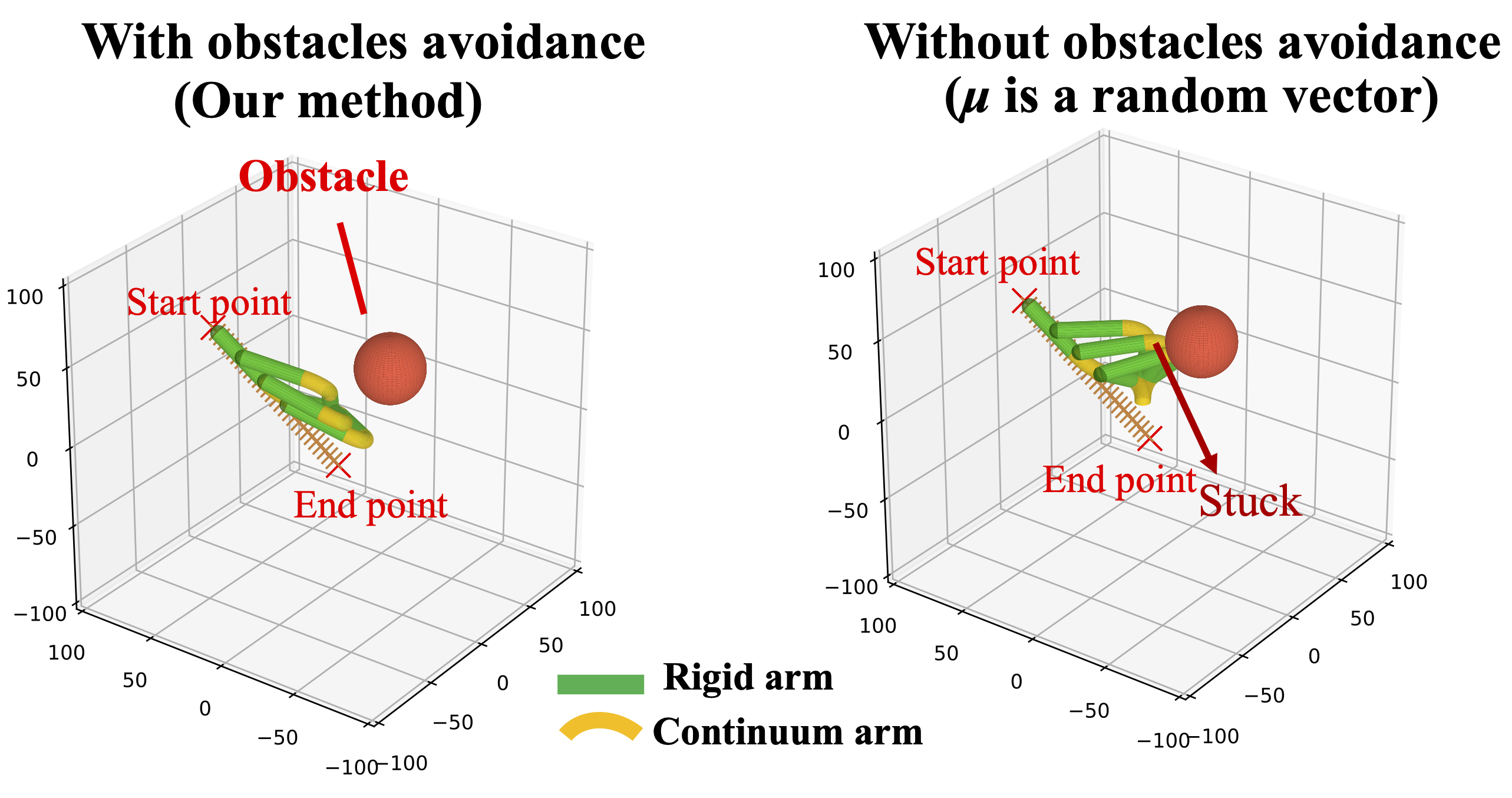}
    \caption{Motion planning comparison with obstacle avoidance planner and planner from \cite{rrt_continuum_compare}.}
    \label{One_obstacle}
    
\end{figure}
\begin{figure*}[t]
    \vspace{0.055in}
    \centering
    \includegraphics[width=\textwidth]{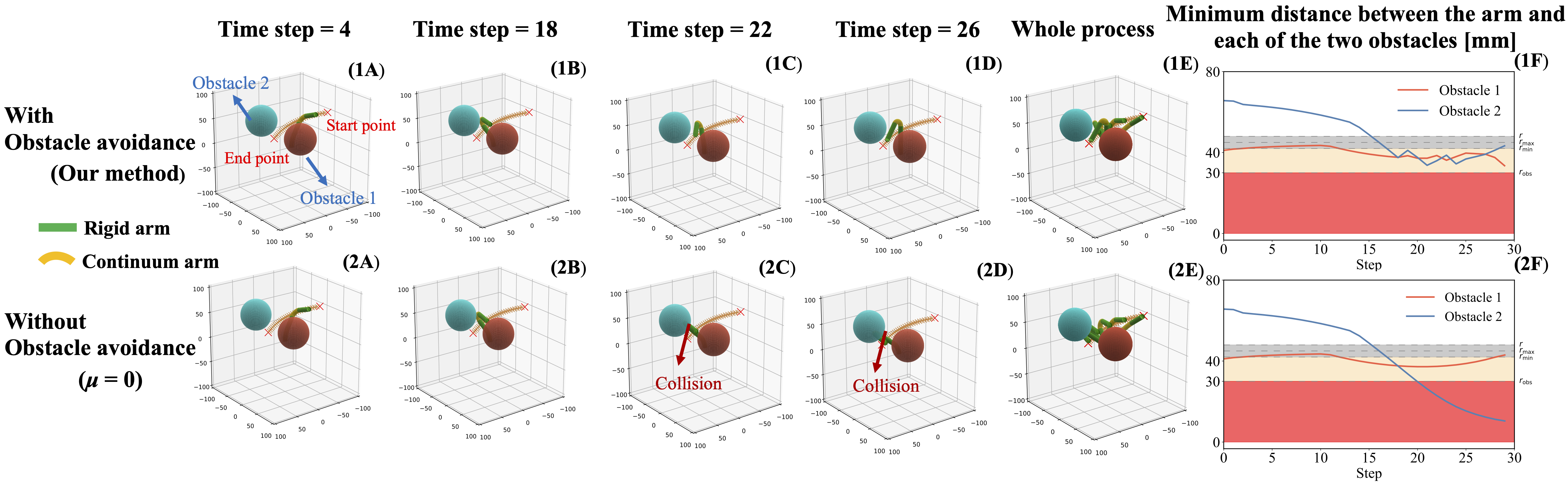}
    \caption{
        Motion planning comparison in an environment with two obstacles at each time step
        (1A) to (1E) shows the results with the obstacle avoidance planner,
        (2A) to (2E) shows the results without the obstacle avoidance planner,
        (1F) displays the minimum distance between the manipulator and the two obstacles with obstacle avoidance planners and (2F) without obstacle avoidance planners.}
    \label{TwoObstacle}
    \vspace{-0.22in}
\end{figure*}

In this section, we simulate the motion in environments with one obstacle and with two obstacles. In the first part, the simulation for the single obstacle case is shown in Fig.\ref{One_obstacle}. The motion planning of the continuum-rigid manipulator with our method and method from \cite{rrt_continuum_compare}. 
The initial configuration is $[\frac{\pi}{3}, \pi, \frac{2\pi}{5}, \frac{\pi}{3}]$, with a startpoint at $[-50, 44, 71]$ and an endpoint at $[-55, -45, 15]$. The obstacle is positioned at $[0, -40, 50]$ with a radius of 20 mm. The search range is 180mm$\times$180mm$\times$90mm. After completing path pruning and smoothing with B-spline curves, 30 points are interpolated to obtain the end-effector trajectory. 
The $r$, $r_{max}$ and $r_{min}$ are set into 38 mm, 35 mm, and 32 mm, and the $k$ in escape velocity for the closest point is set into 6.

Fig.\ref{One_obstacle} illustrates the scenario without obstacle avoidance, using the method described in \cite{rrt_continuum_compare}, where the vector $\boldsymbol{\mu}$ in Eq.\ref{weight_jacob} is a small random vector that slightly deviates from the current configuration $\boldsymbol{q}$. 
In contrast, our obstacle avoidance planner allows the manipulator to proactively adjust its posture, avoiding the obstacle before getting too close. 
However, the method in \cite{rrt_continuum_compare} continues to approach the obstacle without adjusting its posture and eventually gets stuck when the manipulator nears the obstacle. 
This is due to the low sampling efficiency of the random $boldsymbol{\mu}$ when mapping from the four-dimensional C-space to the three-dimensional W-space, causing repeated failures in the collision check for the new configuration.

Next, in the second part, the simulation for motion planning with and without obstacle avoidance in the environment with two obstacles is shown in Fig.\ref{TwoObstacle}(1A) to (1E) and (2A) to (2E). 
The initial configuration and startpoint are the same as in the one-obstacle scenario discussed in the first part, with the endpoint at $[50, 10, 30]$. 
The obstacles are placed at $[10, 40, 30]$ and $[40, -40, 50]$, each with a radius of 30 mm. 
Additionally, 30 points are interpolated to obtain the end-effector trajectory. The $r$, $r_{max}$ and $r_{min}$ are set into 48 mm, 45 mm, and 42 mm, and the $k$ in escape velocity is set into 6. 

Fig.\ref{TwoObstacle}(1F) and (2F) show the minimum distance between the manipulator and each of the two obstacles for both methods. 
In the case without obstacle avoidance, the parameter $\boldsymbol{\mu}$ in Eq. \ref{weight_jacob} is set to 0. 
Initially, the manipulator moves away from obstacle 1 and stays beyond the threshold distance $r$ from obstacle 2. 
According to our Algorithm \ref{S_RRT_motion_planning}, when an obstacle is too far or not in the manipulator's direction, the obstacle avoidance planner does not activate. 
Thus, at first, both planners behave similarly. However, after several time steps, the manipulator approaches both obstacles, and the closest point is no longer the end-effector. 
The obstacle avoidance planner then adjusts the manipulator's posture to avoid the obstacles, while the planner without obstacle avoidance continues toward the obstacles, eventually resulting in a collision.
\subsection{Comparison of Computation Time Across Different Methods}
In this section, we compare the computation time of our method with other methods. The results are shown in Table \ref{computation_time}. 
The environments, as previously described, consist of Environment 1 with a single obstacle and Environment 2 with two obstacles, with all settings kept consistent.

In the path planning step, represented by S-$\text{RRT}^*$, the focus is on finding a feasible path in W-space for the end-effector.
Motion planning refers to the time taken to compute inverse kinematics. 
For comparison, the optimization-based inverse kinematics method \cite{Optimization_iik} shares the same path planning as our method. 
The C-space $\text{RRT}^*$ directly searches for a feasible path in C-space, skipping a separate path planning step. 
Collision detection is done by calculating the minimum distance between the manipulator and obstacles, as detailed in the methodology. 
We also calculate the variance for C-space $\text{RRT}^*$. All experiments were conducted 50 times, and the reported times represent the average computation cost.
Our results show that IIK-based motion planning, which includes both path planning and motion planning, is more time-efficient than using the $\text{RRT}^*$ algorithm directly in C-space. 
Additionally, the computation time variance for C-space $\text{RRT}^*$ is significantly higher. This is because representing obstacles in three-dimensional space (W-space) is more intuitive and efficient compared to four-dimensional space (C-space), allowing $\text{RRT}^*$ to perform better in W-space.
On the other hand, the IIK algorithm is highly stable, maintaining a near-constant solving time at each time step. 
Our method generates one IIK solution in approximately 0.11s, and with a constant time step of 30, the total computation times in the two environments—2.902s and 2.864s—demonstrate its stability.
Compared to other similar methods, such as optimization-based IIK \cite{Optimization_iik}, our method does not require optimization techniques to find the minimum value, which involves time-consuming nonlinear programming. 
Instead, our approach avoids gradient-based searches and directly computes the closest point without iterative sampling of control points, as done in previous methods. 
As shown in Table \ref{computation_time}, our method is approximately 1.5 to 2 times faster. These results demonstrate that our method is not only faster in computation time but also highly stable.

\begin{table}[t]
    \centering
    \scriptsize 
    \caption{COMPUTATION TIME OF DIFFERENT METHODS}
    \label{computation_time}
    \setlength{\tabcolsep}{4pt} 
    \renewcommand{\arraystretch}{1} 
    \begin{tabular}{|c|c|c|c|c|}
        \hline
        \multirow{2}{*}{\backslashbox{\vspace{-1em} \hspace{1em}Method}{\vspace{0.4em} \hspace{0.2em}Env. and Step}} & \multicolumn{2}{c|}{1} & \multicolumn{2}{c|}{2} \\
        \cline{2-5}
        & \makecell{Path\\Planning} & \makecell{Motion\\Planning} & \makecell{Path\\Planning} & \makecell{Motion\\Planning} \\
        \hline
        Our method [s] & \multirow{2}{*}{0.092} & \textbf{2.902} & \multirow{2}{*}{0.141 } & \textbf{2.864}  \\
        \cline{1-1} \cline{3-3} \cline{5-5}
        Optimization-based IIK [s] \cite{Optimization_iik} &  & 5.665 &  & 4.991 \\
        \hline
        C-space $\text{RRT}^*$ [s] (Variance)  & \multicolumn{2}{c|}{87.262 (13935)} & \multicolumn{2}{c|}{19.805 (1046)} \\
        \hline
    \end{tabular}
    \vspace{-0.2in}
    \end{table}

\section{Conclusion and Future Work}
In this work, we proposed an efficient autonomous motion planner for the continuum-rigid manipulator.
We extended the obstacle avoidance technique originally used for rigid arms to continuum arms by analyzing the geometric properties of the continuum arm. By integrating this with the S-$\text{RRT}^*$ algorithm, we achieved autonomous motion planning for continuum-soft arms. 
The simulation results show that our method can generate appropriate joint configurations without obstacle collisions in complex environments, while also demonstrating superior computation time. 
In future work, we plan to extend this technique to dynamic environments and conduct real-world experiments.

\bibliographystyle{IEEEtran}

\bibliography{references}

\end{document}